\documentclass{article}

     \usepackage[nonatbib, final]{neurips_2020}

\usepackage[utf8]{inputenc} 
\usepackage[T1]{fontenc}    
\usepackage{hyperref}       
\usepackage{url}            
\usepackage{booktabs}       
\usepackage{amsfonts}       
\usepackage{nicefrac}       
\usepackage{microtype}      
\usepackage[dvipsnames]{xcolor}
\usepackage{tikz}  
\usepackage{multirow}

\hypersetup{draft}

\title{Deep Reservoir Networks with Learned Hidden Reservoir Weights using Direct Feedback Alignment}

\author{%
    Matthew S. Evanusa \thanks{Corresponding author. Code available at: https://github.com/Symbiomancer} \\
    Department of Computer Science\\
    University of Maryland\\
    College Park, MD \\
    \texttt{mevanusa@umd.edu}
    \And
    Cornelia Ferm\"uller\\
    Institute for Adv. Computer Studies\\
    University of Maryland\\
    College Park, MD\\
    \And
    Yiannis Aloimonos\\
    Department of Computer Science\\
    University of Maryland\\
    College Park, MD\\

}

\begin{document}

\maketitle

\begin{abstract}
 Deep Reservoir Computing has emerged as a new paradigm for deep learning, which is based around the reservoir computing principle of maintaining random pools of neurons combined with hierarchical deep learning. The reservoir paradigm reflects and respects the high degree of recurrence in biological brains, and the role that neuronal dynamics play in learning. However, one issue hampering deep reservoir network development is that one cannot backpropagate through the reservoir layers.  Recent deep reservoir architectures do not learn hidden or hierarchical representations in the same manner as deep artificial neural networks, but rather concatenate all hidden reservoirs together to perform traditional regression.  Here we present a novel Deep Reservoir Network for time series prediction and classification that learns through the non-differentiable hidden reservoir layers using a biologically-inspired backpropagation alternative called Direct Feedback Alignment, which resembles global dopamine signal broadcasting in the brain. We demonstrate its efficacy on two real world multidimensional time series datasets. 
 
\end{abstract}

\section{Introduction}



\subsection{Backpropagation Struggles with Temporal Data}

While much progress has been made towards optimizing feed-forward networks with backpropagation \cite{rumelhart1986learning}, the workhorse for deep learning weight updates since its invention in the 1980s, research has struggled to expand this synchronous, time-less network structure to networks that are capable of learning temporal sequences.  Effectiveness aside, while there are some voices in the community that backpropagation \emph{can} be biologically plausible \cite{lansdell2019learning}, the available evidence seems to indicate that the brain, while potentially performing an operation that reduces error according to some metric, cannot in fact backpropagate errors due to the synaptic structure and strict rules about symmetrical weight updates \cite{lillicrap2016random}.   
Because feed-forward networks contain no memory, one way to use feed-forward networks to encode temporal data is to remove the temporal component entirely, and flatten the entire "series" into one long vector.  Alternatively, the current mode of training gradient-based recurrent neural networks (RNNs) is to directly apply the backpropagation techniques optimized for feed-forward non-temporal networks, to cyclical networks, in the form of RNNs or Long Short Term Memory Networks (LSTMs) \cite{hochreiter1997long}.  These networks can be difficult to train, may require large datasets, and suffer from issues of vanishing gradients and a move away from biological plausibility, although LSTMs aim to address the vanishing gradient concerns.  
More recent variants of LSTMs do away entirely with components in an attempt to simplify the architecture in the Gated Recurrent Unit (GRU) \cite{cho2014learning}.  Other recent branches of backpropagation-led time series learning - Transformer architectures - have eschewed entirely with the recurrent network structure \cite{vaswani2017attention} and go back to learning "time" as a concatenated vector.  Even with these supercharged LSTM, GRU, and Transformer architectures, time series learning remains an open challenge.  Here, we encode time directly in the dynamics of simplified neurons with random connectivity, directly attempting to encode  the dynamical system of the real world in the dynamical system of the network - with the hopes of gleaning insights into how the brain does this.

\subsection{Reservoir Computing}
\label{sec::RC}

Reservoir computing \cite{lukovsevivcius2009reservoir} is a paradigm for recurrent neural network learning that tries to move away and relax the rigid per-time-step gradient updates of LSTMs.  In the reservoir paradigm, there are three components: an input weight matrix (generally untrained) that takes in the input and expands it into a higher dimensional "reservoir".  This second component, the "reservoir", contains random feedback connections, with the maximum eigenvalue set to be less than 1, that bounces around the combined input and recurrent activity for a set time period (while the activity is required to continuously "die out", in order to wash out old inputs \cite{lukovsevivcius2009reservoir}).  The reservoir keeps a history of the expanded temporal input as the different time steps are fed in.  Third, a "readout" mechanism looks at the patterns of activity of the reservoir and performs some learning, which can be a regression or classification task.  For this paper, we show the results of a classification task, although chaotic time series are of high interest to the reservoir community \cite{pathak2018model}.  


\subsection{Current Deep Reservoir Computing Frameworks Do Not Allow for Error-Based Hidden Temporal Representation Learning}

It is well understood that the deep layers of the brain allow for increasingly more complex feature extraction, across multiple domain input types \cite{atencio2009hierarchical}.  In the same way that deep ANNs leveraged deeper networks for more complex feature extraction, Deep ESNs \cite{gallicchio2017deep} attempt to leverage the same hierarchy to extract temporal features from input time series.  However, the current state of deep reservoirs does not learn hidden representations in the same manner that deep learning of ANNs does, for the simple reason that one cannot backpropagate gradients through the reservoir-layers.  Unsupervised learning of these connections does exist in the form of PCA \cite{ma2017deep}, although this does not allow for the same kind of powerful prediction-based error learning as in the hidden layers of deep ANNs.  Here, we take one step towards a deep reservoir that can learn hidden representations with an error generated by the prediction, via direct feedback alignment.  This affords the capability of learning complex \emph{temporal} hidden representations from time series data. 

\section{Contributions}

Our contributions are as follows: 

\begin{itemize}
\item Introduce a novel Deep Reservoir Computer (Deep ESN), with novel trained hidden weights using Direct Feedback Alignment
\item Demonstrate that Direct Feedback Alignment can induce learning through extremely non-differentiable jumps, here over randomly connected neuron pools
\end{itemize}

\vspace{-.5cm}



\begin{figure}[htbp]
\begin{tikzpicture}
  \node (img1) {\includegraphics[width=\textwidth,height=\textheight,keepaspectratio]{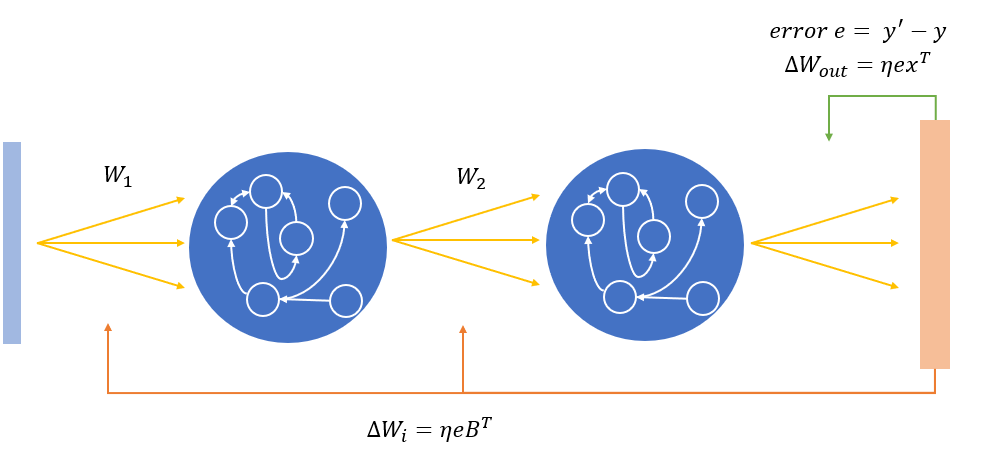}};
\end{tikzpicture}

\caption[Deep]{A schematic of the Deep Reservoir Computing with Learned Hidden Representations using Feedback Alignment framework.  The hidden layer representations, stored as the $W_{in}$ input matrices to the reservoir-layers, are learned through the direct feedback alignment learning mechanism \cite{nokland2016direct}. The error for the last layer is directly fed back to the previous layers through a random matrix $B$ into the corresponding $W_{in}$ weight matrix for that reservoir-layer, and the network \emph{learns to align} with this random matrix.  The weights to the output from the last reservoir layer are computed using the standard delta rule for single-layer ANNs, where $x$ is the activity vector of the last reservoir.  The yellow lines indicate learned connection weights.  The blue circles are the randomly connected reservoir for that layer (while shown with identical connectivity here, they are not necessarily identical). $B$ is a random feedback matrix per \cite{lillicrap2016random, nokland2016direct} with appropriate dimensions, $\eta$ is the learning rate. Shown here are two reservoir-layers, although in our experiments we run with deeper networks effectively. }
\label{fig:Deep}
\end{figure}

\section{Proposed Architecture}

Reservoir networks consists of the rate encoding variant known as Echo State Networks (ESN) \cite{jaeger2001echo}, and a spike (event) based variant, known as Liquid State Machines (LSM) \cite{maass2011liquid}, discovered independently. Here we adopt the ESN approach, although transferring to a LSM is in development.  Each reservoir updates according to the following formula of a simplified reservoir taken from \cite{tong2018reservoir}:


\begin{equation}
    x_t = (1-\alpha)x_{t-1} + \alpha f(W_{in}u_t +  W_{rec}x_{t-1})
\end{equation}

\begin{equation}
    y_t = W_{out} x_t,
\end{equation}

where $y_t$ is the output vector at time $t$, $x_t$ is the $N$-dimensional vector of the states of each reservoir neuron at time $t$, $u_t$ is the input vector at time $t$, $\alpha$ is the leak rate in the interval (0,1), $f(\cdot)$ is a saturating non-linear activation function (here we use a sigmoid), and $W_{in}, W_{rec}, W_{out}$ are the input-to-reservoir, reservoir-to-reservoir, and reservoir-to-output weight matrices, respectively.  If the input vector has length $A$, and the output vector has length $B$, and the reservoir vector size is $N$, then the dimensions of $W_{in}, W_{rec}, W_{out}$ are $N\times A, N \times N$, and $B \times N$, respectively, 

\subsection{Feedback Alignment and Direct Feedback Alignment}

Feedback alignment (FA) has been shown to be a simple yet powerful tool in the quest to move towards more local-update and biologically inspired variants of backpropagation, while at the same time, achieving the same stated end goal of reduction of prediction error for some task.  In the original paper \cite{lillicrap2016random}, the authors point out a few of the major weaknesses of relating backpropagation to biological processes: the fact that these updates must occur in synchronous lock-step, that the synaptic updates must coordinate across multiple neuron layers in a chain to deliver the correct gradient update, and that these updates require weight updates that are symmetric with the output of the neuron.  Feedback alignment attempts to tackle the issue of symmetric weight updates by showing one can simply use a random feedback matrix for the gradient updates, completely unrelated to the feed-forward activity, and still wind up with weight changes within 90 degrees of the true gradient update.  Direct Feedback Alignment \cite{nokland2016direct} takes this idea one step further, and does away with the locality requirements of FA.  There, it was shown that not only can this matrix be random, but it need not come from the layer above: all the updates can be "projected" from the output layer.  This addition throws FA and DFA much closer towards biological theories about reward-modulated STDP with dopamine projections \cite{legenstein2008learning, fremaux2016neuromodulated}, and other neuromodulators, with the error signal $e$ acting as the globally-generated neuromodulator.  

\subsection{Training the Deep ESN with Direct Feedback Alignment}
\label{sec::DFA-ESN}

For the Deep ESN formulation \cite{gallicchio2017deep}, reservoirs are stacked in layers (which we refer to as "reservoir-layers"), analogous to how perceptrons are stacked for deep ANNs.  Each sub-reservoir is given its own unique weight matrices. The reservoir equations in (1) and (2) remain the same for all hidden reservoir-layers but there are two differences.  First the size of $x_t$: for the input reservoir-layer, $x_t$ is the size of the input vector at time $t$, whereas for "hidden" reservoirs, $x_t$ is the size of the preceding reservoir-layer.  This also is analogous to deep ANNs and perceptrons, as each layer takes in as input the output from the last layer.  Second, in the deep formulation, there is only a single $W_{out}$ matrix at the terminal reservoir-layer.  The $W_{in}$ of each subsequent non-terminal reservoir-layer serves as the $W_{out}$ for the previous reservoir-layer. At regularly sampled time intervals, the correct label $y$ is compared against the output of the network, $y'$.  The last layer of the network is trained using the traditional delta rule for perceptron learning (i.e., the last layer of a deep ANN) which reduces error via gradient descent; the $W_{in}$ matrices of each preceding reservoir-layer are trained using the direct feedback alignment algorithm.  The hypothesis is that the error from the last output layer will "align" these intermediary reservoir-layer weights to reduce the output error, \emph{even though} the "true" gradient is not propagated backwards through all time steps. We illustrate the architecture in figure \ref{fig:Deep} for a simple two-reservoir-layer network. We refer to the deep ESN trained with DFA as DFA-DeepESN. 

The values of the input weight matrix leading into reservoir i, $W_i$, are updated at each update step using the direct feedback alignment update: 

\begin{equation}
    \Delta W_i = \eta  e B_{i}^{T} 
\end{equation}

where $\eta$ is the learning rate, $e$ is the last-layer error, and $B^T$ is the transpose of the random matrix of appropriate dimension (See Fig. \ref{fig:Deep}).

\section{Results}

We tested our network on several challenging real-world high-dimensional input time-series classification datasets: BasicMotion and ERing, which are taken from the UEA Multivariate Time Series database \cite{bagnall2018uea} and are freely and publicly available online.  The results are shown below.  BasicMotions has 6 input dimensions per time step and 4 outputs corresponding to different movements (walking, running, standing, and playing badminton) taken from a HAR sensor. ERing has 4 input dimensions of a prototype finger sensor and 6 outputs corresponding to the finger.  For all series, the reservoir activity was sampled at regular intervals for readout training to the target label for the given series.  The weight change updates were performed as one large batch at the end of each epoch for both the last layer and the $W_i$ weights. For BasicMotion, reservoirs of size 800 were used.  For the deep networks, 4 reservoir-layers were used for all experiments. We set $\alpha$ to 0.1, weight decay at $10e^-9$, learning rate $\eta$ to 0.01, and learning rate decay at $10e^-7$ per epoch.

\begin{table}[h!]
\begin{center}
\begin{tabular}{ll}
\begin{tabular}{|c|c|c|}
\hline
\textbf{BasicMotions} & \textbf{train} & \textbf{test} \\
\hline
Single-Reservoir & 82.5 & 77.5 \\
\hline
DeepESN w/out DFA & 100 & 97.5\\
\hline
DFA-DeepESN & 100.0 & \textbf{100.0}\\
\hline

\end{tabular}

\begin{tabular}{|c|c|c|}
\hline
\textbf{ERing} & \textbf{train} & \textbf{test} \\
\hline
Single-Reservoir & 68.15  & 73.3 \\
\hline
DeepESN w/out DFA & 93.3 & 78.15\\
\hline
DFA-DeepESN & \textbf{96.7} & \textbf{79.26}\\
\hline

\end{tabular}
\end{tabular}

\end{center}
\caption{Results of the accuracy of variants of the network, either without DFA or non-deep reservoirs, on both the BasicMotions and ERing datasets.  Results are shown as \% correct classification accuracy.}
\end{table}

\vspace{-.5cm}

\section*{Conclusion and Future Work}

Here we propose a novel deep reservoir network with learned hidden weight matrices (DFA-DeepESN), using the Direct Feedback Alignment method.  Our hope is that this work will allow future recurrent networks that have highly non-differentiable components to train effectively towards temporal predictions. Future work will involve analysis of the hidden temporal features themselves, and testing of alternative backpropagation alternatives for the weight training.

\begin{ack}

The authors would like to thank Vaishnavi Patil for her invaluable assistance in this endeavor. This work was supported in part by NSF award DGE-1632976.

\end{ack}

\bibliography{biblio}

\end{document}